\documentclass[aps,pre,reprint,amsmath,amssymb]{revtex4-2}
\usepackage{txfonts}
\usepackage{algorithmic}
\usepackage{algorithm}
\usepackage{graphicx}
\usepackage{bm}
\usepackage{url}
\usepackage{color}
\usepackage{svg}

\newcommand{\mat}[1]{\bm{#1}}
\newcommand{\R}{\mathbb{R}}

\begin{document}

\title{Neuron merging via inverse-activation regression for post-training compression of sigmoid neural networks}

\author{Ao Kuniya and Jun Ohkubo}

\affiliation{Graduate School of Science and Engineering, Saitama University, Sakura, Saitama, 338--8570 Japan}

\begin{abstract}
As neural networks continue to grow in scale, model compression is becoming increasingly important for efficient inference under limited computational resources. Structured pruning methods remove neurons or channels that are estimated to be less important, but the removed units may still contain useful information. From the viewpoint of coarse-graining a trained network, it is valuable to ask which information should be retained when multiple neuronal degrees of freedom are consolidated. In this paper, we discuss cluster-based merging methods for compression of trained neural networks. In addition to a data-free contribution-weighted averaging method, we propose neuron-merging methods in which neuron responses are mapped back to the pre-activation space via the inverse activation function, and the weights and biases of each representative neuron are estimated using the least-squares method. We also examine both a data-assisted strategy with actual training inputs and a data-free strategy using randomly generated inputs. The comparisons provide empirical evidence, in the tested sigmoid networks, that weight information is particularly useful for clustering whereas activation information is useful for representative-neuron reconstruction in the merging process.
\end{abstract}

\maketitle

\section{Introduction}
\label{sec:introduction}

In recent years, research on neural networks has advanced significantly; the importance of machine learning was also highlighted by the 2024 Nobel Prizes in Physics and Chemistry. Furthermore, their practical applications in our daily lives are expanding. However, these neural networks are becoming increasingly complex, containing more and more parameters. Their deployment has expanded across many applications, increasing concern about the computational and energy costs of both training \cite{Strubell2019,Strubell2020} and inference \cite{Fernandez2025,Caravaca2025}. Hence, compressing a trained network is important when memory, computation, or inference latency is limited.

Knowledge distillation \cite{Hinton2015} is one well-known method for efficiently utilizing a trained neural network; see recent review papers \cite{Moslemi2024,Mansourian2025}. It transfers knowledge from a teacher model to a student model, which is often chosen to be smaller or cheaper to evaluate. Note that this process requires additional training. Pruning is another method for compressing neural networks; see recent review papers \cite{Liang2021,He2024,Cheng2024}. Pruned networks can retain useful performance without fine-tuning at moderate compression levels, although fine-tuning often improves accuracy. Generally, weight matrices are dense, but some matrix elements contribute only minimally to the results. Therefore, pruning involves removing neurons or connections between neurons that seem unimportant. Unstructured pruning removes connections between neurons, thereby reducing the number of parameters. Note that the weight matrix elements set to zero due to removal appear irregularly, so the resulting irregular matrices do not necessarily provide speedups without specialized sparse implementations \cite{Han2015}. On the other hand, structured pruning removes neurons and channels rather than connections. Hence, it generates smaller, higher-density weight matrices and reduces the computational cost during inference \cite{Li2017,Molchanov2017}. Of course, structured pruning removes neurons that may still carry information contributing to subsequent layers. This loss becomes particularly important at strong compression levels. Although fine-tuning partially recovers the lost accuracy, it requires additional data and an optimization procedure. In the context of physics, some recent works focus on the pruning and related techniques; for example, see \cite{Acharyya2021,Mori2025,Surkov2026}.

As a simple idea, it is easy to imagine that information from pruned neurons could contribute to effective compression without the additional fine-tuning. Neuron merging offers an alternative to simple pruning. If several neurons exhibit similar input-output relationships, their contributions might be consolidated into a single representative neuron. Some works mainly compress networks by identifying redundant neurons from weights or simple neuron behavior and then pruning or clustering/merging them; data-free pairwise redundancy with a compensating surgery \cite{Srinivas2015}, weight-based clustering/agglomeration \cite{Zhong2018,Wang2020}, and inactive/similar ReLU neurons in a restricted single-input–single-output architecture \cite{Emerald2025}. Especially, Kim \textit{et al.} formulated neuron merging by introducing a scaling matrix that transfers the contributions of removed neurons to the next layer \cite{Kim2020}. As discussed later, the method of Kim \textit{et al.} \cite{Kim2020} is data-free; it reconstructs the output feature maps of the original neural networks better than vanilla pruning only from the weight matrices. However, the merging method in \cite{Kim2020} relies on the algebraic properties of the ReLU functions and does not directly apply to sigmoid networks.

In the present paper, we propose several specific merging algorithms to investigate what types of information are useful for neuron merging. In particular, we examine not only the use of information from weight matrices but also that from activation functions. We first detect clusters of neurons to be merged using the values of the weight matrices or activation functions. In addition to the naive averaging method for the weight vectors, we propose methods that employ the activation function values to estimate the weights and biases of representative neurons via the least-squares method. The proposed reconstruction maps activations to the pre-activation space through the inverse activation function and then performs ridge-regularized least-squares regression; here, we focus on the sigmoid function. Specifically, we implement the framework using the logit function, which is the inverse of the sigmoid function. We numerically evaluate variations of our proposals, including data-assisted and training-data-free versions, using both weight-based and activation values-based clustering, and demonstrate that they outperform structured pruning. Furthermore, we clarify the roles of weight matrices and activation function values in neuron merging.

The remainder of this paper is organized as follows. Section~\ref{sec:background} reviews structured pruning and neuron merging; we will also show numerical results on compression using standard structured pruning. Section~\ref{sec:method} presents the main proposals of the present paper; we propose several merging methods to investigate the roles of information in the weight matrices and the activation function values. Then, we demonstrate the proposals in Sect.~\ref{sec:experiments} and discuss the roles in the merging process. From a practical perspective, the required computational time for the merging process is also discussed. Finally, in Sect.~\ref{sec:conclusion}, we present our conclusions and discuss future research directions.

\section{Background and Naive Demonstration}
\label{sec:background}

\subsection{Structured pruning}
\label{subsec:structured_pruning}

\begin{figure}[tb]
  \centering
  \includegraphics[width=85mm]{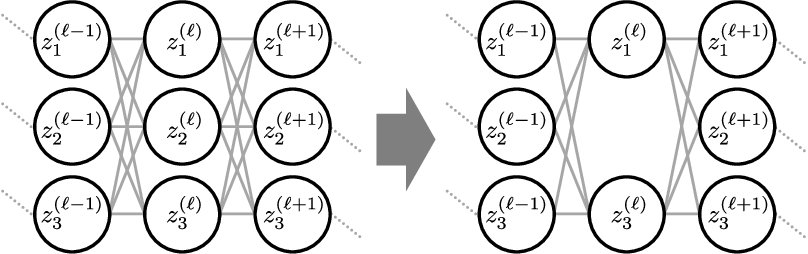}
  \caption{Structured pruning. In unstructured pruning, connections between neurons are removed individually. By contrast, structured pruning removes neurons and all their associated connections, thereby deleting the corresponding rows or columns from the weight matrices. Hence, the overall sizes of the weight matrices are reduced.}
  \label{fig:pruning_overview}
\end{figure}

We here briefly review the conventional pruning methods. Figure~\ref{fig:pruning_overview} shows the concept of structured pruning. In unstructured pruning, connections between neurons are removed individually. As a result, only some elements of the weight matrices become zero; while this yields sparse matrices, the sizes of the weight matrices themselves do not decrease. By contrast, structured pruning removes neurons according to their estimated importance. Since we delete the corresponding rows or columns from the weight matrices, the overall sizes of the weight matrices are reduced, and the resulting network retains dense matrix operations. Hence, it reduces practical inference costs \cite{Li2017}. 

While magnitude-based methods naively use weights to rank units, there are several approaches to ranking. Li \textit{et al.} used the $\ell_1$ norm to identify filters for structured removal \cite{Li2017}, while Molchanov \textit{et al.} evaluated weight-magnitude criteria and proposed a first-order Taylor criterion based on the product of activation and loss gradient \cite{Molchanov2017}. Data-dependent criteria include the mean absolute activation and the Fisher score based on a squared activation--gradient signal \cite{Theis2018}. He \textit{et al.} combined sparse channel selection with least-squares reconstruction of layer responses \cite{He2017}. 

As mentioned above, while these structured pruning methods reduce memory and computational complexity, they may lose information carried by the removed units.

\subsection{Demonstration of structured pruning}
\label{subsec:demo_pruning}

\begin{table}[tb]
  \caption{Training settings.}
  \label{tab:experimental_setting}
  \centering
  \begin{tabular}{ll}
    \hline\hline
    Item & Setting \\
    \hline
    Model & 784--1024--1024--1024--10 \\
    Hidden activation \,\,\,& Sigmoid function\\
    Optimizer & Adadelta with learning rate $1.0$ \\
    Epochs & 20 \\
    Batch size & 128 \\
    \hline\hline
  \end{tabular}
\end{table}

Here, we present the structured-pruning results on the MNIST dataset for handwritten digit recognition \cite{Lecun1998}. Each image is $28 \times 28$, and we standardize and flatten it into a 784-dimensional input. Table~\ref{tab:experimental_setting} shows the settings for the neural network training. The neural network has three sigmoid hidden layers of width 1024 and a ten-dimensional linear output layer. We train ten independent neural networks with different random seeds. After constructing neural networks using the MNIST training data, we prune them and evaluate their classification accuracy on 10,000 test data points. In the pruning procedure, the same number of neurons is removed from each hidden layer. Hence, the widths of the hidden layers after pruning are the same for all of them. Note that we do not perform a fine-tuning procedure in these experiments. The MNIST dataset consists of 60,000 training examples, and we set the batch size to 128. During testing, we employ all 10,000 test examples in the MNIST dataset.

In the current numerical experiments, when the number of neurons in hidden layer $\ell$ is reduced from $d_{\ell}$ to $K_{\ell}$, the \textit{node-retention ratio}, $r_{\mathrm{ret}}$, is defined as follows:
\begin{align}
  r_{\mathrm{ret}}
  = \frac{\sum_{\ell} K_{\ell}}{\sum_{\ell} d_{\ell}},
  \label{eq:node_retention}
\end{align}
where the sums run over the three hidden layers. Hence, a smaller value of $r_{\mathrm{ret}}$ indicates stronger compression.

\begin{figure}[tb]
  \centering
  \includegraphics[width=85mm]{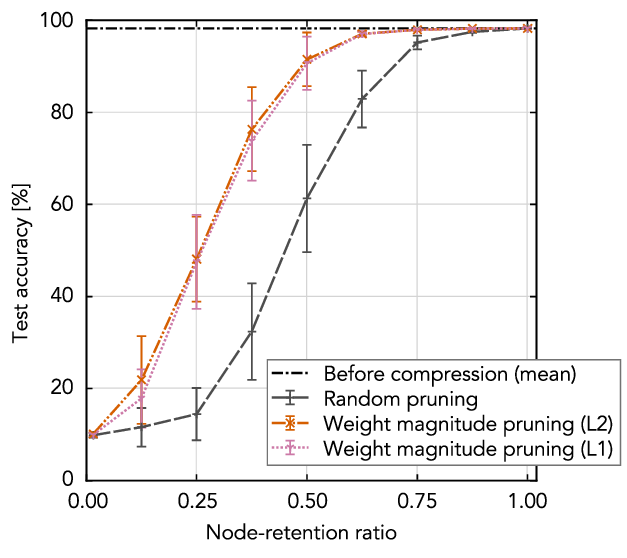}
  \caption{
(Color online) Comparison of compression and accuracy through several structured pruning methods. The horizontal axis shows the node-retention ratio in the hidden layers, and the vertical axis shows the accuracy on the test data. The $\ell_1$- and $\ell_2$-norm weight-magnitude pruning methods are data-free and utilize only the weight information. Symbols and error bars denote the mean and standard deviation over ten independently trained networks.}
  \label{fig:results_pruning}
\end{figure}

Figure~\ref{fig:results_pruning} shows the numerical results. Here, we evaluate three structured pruning methods: Random pruning, $\ell_1$- and $\ell_2$-norm weight-magnitude pruning \cite{Li2017,Molchanov2017}. The numerical results show that while accuracy is maintained to some extent even with a slight compression, it rapidly decreases at a certain point. Note that the $\ell_1$- and $\ell_2$-norm weight-magnitude pruning methods are data-free and utilize weight information.

\subsection{Neuron merging with ReLU function}
\label{subsec:neuron_merging}

As mentioned in Sect.~\ref{sec:introduction}, the neuron merging method by Kim \textit{et al.} makes efficient use of information from deleted neurons \cite{Kim2020}. In the neuron merging method, one starts by decomposing the original weights into two matrices. One of them corresponds to the new weights for the current layer, and the other is called a scaling matrix that yields a linear combination of neurons. Hence, the information from pruned neurons can be utilized efficiently. Numerical demonstrations in Ref.~\cite{Kim2020} exhibited that the merged models without fine-tuning work better than the naive pruned models.

Note that Kim \textit{et al.} discussed only a certain specific situation \cite{Kim2020}. That is, in their derivation, the scaling matrix is moved outside the activation function and incorporated into the next layer; this step is valid for the rectified linear unit (ReLU) function when the scaling matrix satisfies specific nonnegativity and sparsity conditions. 

In general, one would expect that utilizing data in the merging process could yield a higher-performance merged network. Since the neuron merging method in \cite{Kim2020} is data-free, other approaches are needed to examine the role of data and activation function values. Hence, we propose an approach that employs the sigmoid function rather than the ReLU function in \cite{Kim2020}.


\section{Proposed Methods}
\label{sec:method}

To investigate the role of input-dependent activation information, we propose the following merging procedures. First, we use clustering to select which neurons to merge. Here, we introduce two approaches: a data-free method and a data-dependent method that exploits information about activation function values. Furthermore, in the merging step, in addition to a simple data-free method based on the average of weights, we propose a least-squares approach that uses the inverse of the sigmoid function and activation function values derived from the data. By combining these methods, it is possible to investigate the role of data in each step, i.e., neuron selection and merging. These procedures are described in detail below.

Note that the data-dependent methods referred to here rely solely on input data; output label data is not required.

\subsection{Problem formulation}
\label{subsec:problem_formulation}

Consider a feedforward network whose hidden activation function $\phi$ is invertible on the range of interest. Focusing on the $n$-th input vector $\bm{x}_n$, we denote the activation function value of neuron $j$ on layer $\ell$ as $z_j^{(\ell)}(\bm{x}_n)$. Let $W_{ij}^{(\ell)}$ be the weight parameter between neuron $i$ on layer $\ell-1$ and neuron $j$ on layer $\ell$. Then, we have
\begin{align}
  z_j^{(\ell)}\left(\bm{x}_n \right)
  = \phi\left(
      \sum_{i} W_{ij}^{(\ell)} z_i^{(\ell-1)}\left(\bm{x}_n \right) + b_j^{(\ell)}
    \right),
  \label{eq:layer_activation}
\end{align}
where $b_j^{(\ell)}$ is the corresponding bias term.

\begin{figure}[t]
  \centering
  \includegraphics[width=85mm]{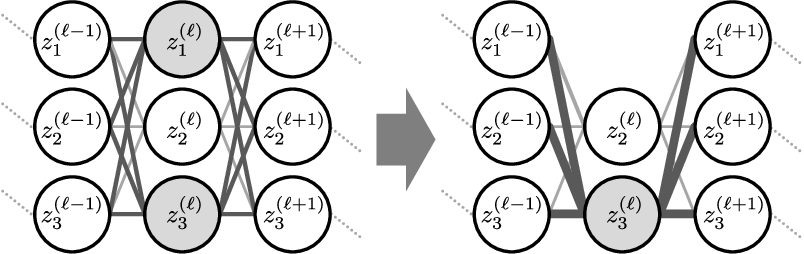}
  \caption{Conceptual illustration of neuron merging. Here, similar neurons $1$ and $3$ are replaced by one representative neuron depicted by the shaded circle. The representative incoming and outgoing parameters are reconstructed.}
  \label{fig:merge_overview}
\end{figure}

Figure~\ref{fig:merge_overview} illustrates the structural change produced by the neuron merging procedure. In Fig.~\ref{fig:merge_overview}, only two neurons are merged into a single representative neuron, but as will be explained later, clustering is used to merge multiple neurons. The representative-neuron parameters are constructed. As explained below, we introduce both data-free and data-dependent methods for clustering and representative neuron parameters.

\subsection{Neuron clustering}
\label{subsec:clustering}

\subsubsection{Data-free clustering approach (C-weight)}

Let layer $\ell$ contain $d_{\ell}$ neurons. We divide the neurons into $K_{\ell} < d_{\ell}$ disjoint clusters $\mathcal{C}^{(\ell)}=\Big\{ C_1^{(\ell)},\ldots,C_{K_{\ell}}^{(\ell)} \Big\}$ and replace neurons in each cluster by one representative neuron. The resulting layer has $K_{\ell}$ neurons and remains a standard dense layer. 

The first clustering method is data-free. Hereinafter, we will refer to this clustering method as \textit{C-weight}. Based on the results of several preliminary experiments, we adopt a feature vector that concatenates the input weights, bias, and output weights. That is, we set the feature vector for neuron $j$ on layer $\ell$ as
\begin{align}
  \bm{q}_{j, \textrm{C-weight}}^{(\ell)}
  = \left[
      W_{1j}^{(\ell)}, \dots, W_{d_{\ell-1}j}^{(\ell)}, b_j^{(\ell)}, W_{j1}^{(\ell+1)}, \dots, W_{j d_{\ell+1}}^{(\ell+1)}
    \right].
  \label{eq:weight_feature}
\end{align}
Using these feature vectors, we standardize the collected feature vectors on each feature coordinate and then apply the conventional $k$-means algorithm in \verb|scikit-learn| \cite{scikit-learn}.

Note that including output weight parameters allows effective neuron clustering; in preliminary experiments, we found that output weight parameters are empirically useful for detecting similarities among neurons, as well as input weight parameters.

\subsubsection{Data-dependent clustering approach (C-activation)}

The second clustering method is data-dependent, i.e., we employ the activation function values. We will refer it as \textit{C-activation}. Here, we feed a portion of the training data into the trained neural networks and use the resulting activation function values to construct a feature vector. Let $N_{\mathrm{comp}}$ denote the number of data points used to construct the feature vector. Then, the feature vector of neuron $j$ on layer $\ell$ is represented as
\begin{align}
  \bm{q}_{j, \textrm{C-activation}}^{(\ell)}
  = \left[
      z_j^{(\ell)}(\bm{x}_1), \dots,
      z_j^{(\ell)}(\bm{x}_{N_{\mathrm{comp}}})
    \right],
  \label{eq:activation_feature}
\end{align}
where the data points are selected uniformly at random.

Except for the construction of the feature vectors, the other procedures are the same as those in C-weight, and the neurons to be merged are selected through the conventional $k$-means clustering after standardization.

\subsection{Weight merging}
\label{subsec:weight_merging}

\subsubsection{Data-free merging approach (M-av)}

As for the data-free approach, one may adopt simple averaging as the data-free approach. For example, the incoming-weight parameters, $\Big\{\widetilde{W}_{1k}^{(\ell)}, \dots, \widetilde{W}_{d_{\ell-1} k}^{(\ell)} \Big\}$, for the representative neuron $k$ of cluster $C_k^{(\ell)}$ become as follows:
\begin{align}
\widetilde{W}_{ik}^{(\ell)} = \frac{1}{\big\lvert C_{k}^{(\ell)} \big\rvert}  \sum_{j \in C_{k}^{(\ell)}} W_{ij}^{(\ell)}.
\end{align}

However, we here take a different approach than the uniform averaging. This is because preliminary experiments showed that the following approach yielded better results. We first assign the following contribution score $s_j^{(\ell)}$ to neuron $j$ on layer $\ell$:
\begin{align}
  s_j^{(\ell)}
  = \gamma_{\mathrm{in}}\|\bm{w}_{j,\mathrm{in}}^{(\ell)}\|_2
  + \gamma_b \left\lvert b_j^{(\ell)} \right\rvert
  + \gamma_{\mathrm{out}}\|\bm{w}_{j,\mathrm{out}}^{(\ell+1)}\|_2,
  \label{eq:parameter_score}
\end{align}
where 
\begin{align}
\bm{w}_{j,\mathrm{in}}^{(\ell)} = \Big[ W_{1j}^{(\ell)}, \dots, W_{d_{\ell-1}j}^{(\ell)}  \Big]
\end{align}
is the incoming-weight vector to neuron $j$, and 
\begin{align}
\bm{w}_{j,\mathrm{out}}^{(\ell+1)} = \Big[ W_{j1}^{(\ell+1)}, \dots, W_{j d_{\ell+1}}^{(\ell+1)}  \Big]
\label{eq:outgoing_vector}
\end{align}
is the outgoing-weight vector from neuron $j$.

Within cluster $C_k^{(\ell)}$, the normalized coefficient is
\begin{align}
  \beta_{kj}^{(\ell)}
  = \frac{s_j^{(\ell)}}{\sum_{r\in C_k^{(\ell)}}s_r^{(\ell)}}.
  \label{eq:parameter_merge_coefficient}
\end{align}
Then, the weight parameters of the representative neuron are given as follows:
\begin{align}
  &\widetilde{\bm{w}}_{k,\mathrm{in}}^{(\ell)}
  = \sum_{j\in C_k^{(\ell)}}
     \beta_{kj}^{(\ell)}\bm{w}_{j,\mathrm{in}}^{(\ell)},
\label{eq:M_av_1}\\
  &
  \widetilde{b}_k^{(\ell)}
  = \sum_{j\in C_k^{(\ell)}}
     \beta_{kj}^{(\ell)}b_j^{(\ell)},
\label{eq:M_av_2}\\
  &\widetilde{\bm{w}}_{k,\mathrm{out}}^{(l+1)}
  = \left\lvert C_k^{(\ell)} \right\rvert
     \sum_{j\in C_k^{(\ell)}}
     \beta_{kj}^{(\ell)}\bm{w}_{j,\mathrm{out}}^{(\ell+1)}.
\label{eq:M_av_3}
\end{align}

We call the merging method based on Eqs.~\eqref{eq:M_av_1}, \eqref{eq:M_av_2}, and \eqref{eq:M_av_3} as \textit{M-av}.

\subsubsection{Data-dependent merging approach (M-logit)}

Here, we discuss a data-dependent approach. In the following discussions and numerical experiments, we employ the sigmoid function,
\begin{align}
\sigma(a) = \frac{1}{1 + \exp(-a)},
\end{align}
as the activation function $\phi(a)$. As a data-dependent approach, we focus on the characteristics of the sigmoid function. For example, suppose that neurons $1$ and $3$ in layer $\ell$ are merged into representative neuron $3$, as in Fig.~\ref{fig:merge_overview}. Then, we seek weight parameters of the representative neuron satisfying the following relation for an input vector $\bm{x}_n$:
\begin{align}
\frac{1}{2} z_{1}^{(\ell)}(\bm{x}_n) + \frac{1}{2} z_3^{(\ell)}(\bm{x}_n)
=
\sigma\left( \sum_{j} \widetilde{W}_{j3}^{(\ell)} z_{j}^{(\ell-1)}(\bm{x}_n) + \widetilde{b}_3^{(\ell)} \right),
\end{align}
where $\Big\{\widetilde{W}_{j3}^{(\ell)}\Big\}$ and $\widetilde{b}_3^{(\ell)}$ are the weight parameters and the bias of the representative neuron, respectively. Since the sigmoid function has the inverse function, and we have
\begin{align}
\sigma^{-1}\left(
\frac{1}{2} z_{1}^{(\ell)}(\bm{x}_n) + \frac{1}{2} z_3^{(\ell)}(\bm{x}_n)
\right) 
= \sum_{j} \widetilde{W}_{j3}^{(\ell)} z_{j}^{(\ell-1)}(\bm{x}_n) + \widetilde{b}_3^{(\ell)}.
\label{eq:example_M_logit}
\end{align}
Note that the left-hand side of Eq.~\eqref{eq:example_M_logit} is calculated using specific numerical values for the input $\bm{x}_n$. Hence, evaluating Eq.~\eqref{eq:example_M_logit} for a large amount of data allows us to estimate the representative incoming weights and bias from the data as a least-squares problem.

From the above example, we construct the following merging procedure. In the following discussion, we denote the inverse function of the sigmoid function as $\operatorname{logit}$, i.e.,
\begin{align}
\operatorname{logit}(a) = \ln \frac{a}{1-a}.
\label{eq:logit}
\end{align}
Then, for cluster $C_k^{(\ell)}$, we define the target response on an input vector $\bm{x}_n$ as a convex combination of the cluster activations:
\begin{align}
  t_{nk}^{(\ell)}
  = \sum_{j\in C_k^{(\ell)}}
    \alpha_{kj}^{(\ell)} z_j^{(\ell)}(\bm{x}_n),
  \qquad
  \sum_{j\in C_k^{(\ell)}}\alpha_{kj}^{(\ell)}=1.
  \label{eq:target_activation}
\end{align}

Although one may use the simple uniform coefficients $\Big\{\alpha_{kj}^{(\ell)}\Big\}$, i.e., $\alpha_{kj}^{(\ell)}=1/\big|C_k^{(\ell)}\big|$, there are other possible settings. Based on several preliminary numerical experiments, we employ the following contribution-aware target based on the outgoing-weight norm:
\begin{align}
  \alpha_{kj}^{(\ell)}
  = \frac{\| \bm{w}_{j,\mathrm{out}}^{(\ell+1)}\|_2}
  {\sum_{r\in C_k^{(\ell)}}\|\bm{w}_{r,\mathrm{out}}^{(\ell+1)}\|_2},
  \label{eq:weighted_target}
\end{align}
where $\bm{w}_{j,\mathrm{out}}^{(\ell+1)}$ is the outgoing-vector defined in Eq.~\eqref{eq:outgoing_vector}. We employ Eq.~\eqref{eq:weighted_target} in the numerical experiments in Sect.~\ref{sec:experiments}.

For practical purposes, the arguments of the logit function must fall within the range from 0 to 1.

In practice, the argument of the logit function in Eq.~\eqref{eq:logit} must fall within the range from $0$ to $1$. Hence, we clip the target for numerical stability. That is, we set
\begin{align}
  \bar{t}_{nk}^{(\ell)}
  &= \operatorname{clip}
     \left(t_{nk}^{(\ell)},\epsilon,1-\epsilon\right),
  \label{eq:target_clip}
\end{align}
and we apply the logit function to the clipped one:
\begin{align}
  u_{nk}^{(\ell)}
  &= \operatorname{logit}\left(\bar{t}_{nk}^{(\ell)}\right)
   = \log\frac{\bar{t}_{nk}^{(\ell)}}{1-\bar{t}_{nk}^{(\ell)}}.
  \label{eq:logit_target}
\end{align}
In the following experiments, we use $\epsilon = 10^{-5}$ in Eq.~\eqref{eq:target_clip}.

The least-squares problem is formulated as follows. Let $X^{(\ell)}\in\R^{N_{\mathrm{comp}}\times(d_{l-1}+1)}$ be the data matrix; this data matrix consists of vectors of the activation function values on layer $\ell-1$ for an input vector $\bm{x}_n$ and an element $1$ corresponding to the bias term. In addition, a matrix $U^{(\ell)}\in\R^{N_{\mathrm{comp}}\times K_l}$ consists of $\{u_{nk}^{(\ell)}\}$ in Eq.~\eqref{eq:logit_target}. Combining the weight parameters and the intercept terms into $\Theta^{\ell}$, as in the conventional least-squares problem, it is enough to solve the following least-squares problem:
\begin{align}
  \widehat{\Theta}^{(\ell)}
  = \arg\min_{\mat{\Theta}} 
    \left\|X^{(\ell)} \Theta - U^{(\ell)}\right\|_F^2.
  \label{eq:representative_regression}
\end{align}
Because the regression result minimizes error in the transformed pre-activation space, Eq.~\eqref{eq:representative_regression} is a linear surrogate for direct activation-space reconstruction rather than an exact minimizer of activation mean squared error of the original task.

As for the outgoing-weight parameters, there is no need to consider the activation function values. Hence, we employ the following simple averaging. After reconstructing representative neuron $k$, we merge the outgoing weights of its cluster as 
\begin{align}
  \widetilde{W}_{ki}^{(\ell+1)}
  = \sum_{j\in C_k^{(\ell)}} W_{ji}^{(\ell+1)}.
  \label{eq:outgoing_merge}
\end{align}
This expression is accurate when neurons in the same cluster have similar responses. 

Of course, it is also possible to set the outgoing-weight parameters using Eq.~\eqref{eq:M_av_3}, and this approach could yield better performance. However, to focus only on the role of the activation function values, we avoid the use of the merging method based on Eq.~\eqref{eq:M_av_3} in the M-logit approach.

\subsection{Comments on the merging algorithms}
\label{subsec:layerwise_algorithm}

The clustering, incoming-weight merging, and outgoing-weight merging are applied sequentially to each hidden layer from $\ell = 1$ to the final one. In data-dependent cases, we first evaluate the activation function values in the first layer using the input dataset; those values are used to perform clustering and merging. Next, we feed the same input dataset into the merged neural network to evaluate the activation function values in the next layer, and we repeat this process for all hidden layers.

The proposed methods have a one-time compression cost for activation collection, clustering, and regression. Its inference graph, however, consists only of the compressed dense layers and the original activation functions.  Therefore, it does not introduce an additional inference-time relative to a network with the same compressed architecture.

\section{Numerical Experiments and Discussions}
\label{sec:experiments}

\subsection{Experiments on MNIST}
\label{subsec:results_MNIST}

We first use the MNIST dataset \cite{Lecun1998}. Each image is standardized and flattened into a 784-dimensional input. The trained neural network is constructed in the same way with Sect.~\ref{subsec:demo_pruning}; we train ten independent neural networks with different random seeds.

We evaluate the combination of the two clustering methods for selecting neurons to merge, i.e., C-weight and C-activation in Sect.~\ref{subsec:clustering}, and the two methods for merging neurons, i.e., M-av and M-logit in Sect.~\ref{subsec:weight_merging}. As for the C-activation and M-logit, we set the number of data points for the C-activation and M-logit, $N_{\mathrm{comp}}$, as $10,000$. The hyperparameters for M-av in Eq.~\eqref{eq:parameter_score} are set to $\gamma_{\mathrm{in}} = \gamma_b = \gamma_{\mathrm{out}} = 1$; we have checked that even if these hyperparameters are altered slightly, the conclusion remains the same.

\subsubsection{Accuracy evaluation}

\begin{figure}[tb]
  \centering
  \includegraphics[width=85mm]{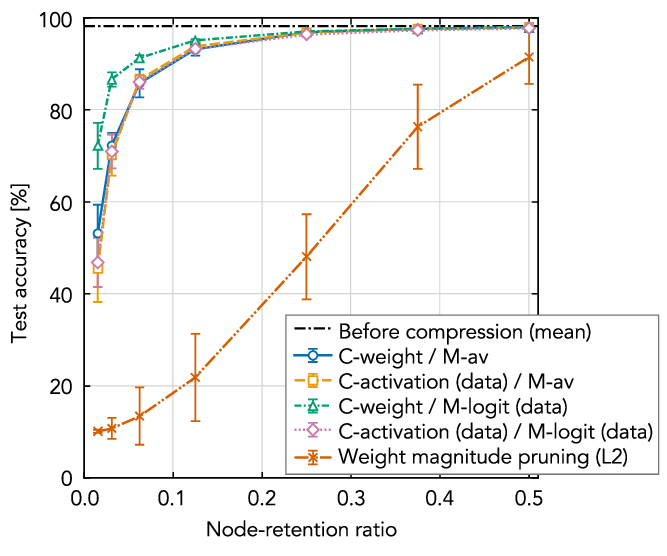}
\caption{(Color online) Test accuracy after neuron merging in the MNIST dataset experiments. The horizontal axis shows the node-retention ratio in the hidden layers, and the vertical axis shows the accuracy on the test data. For reference, this figure also shows the results of ``Weight magnitude pruning (L2)'', as depicted in Fig.~\ref{fig:results_pruning}. Symbols and error bars denote the mean and standard deviation over ten independently trained networks.
}
\label{fig:mnist_results_accuracy}
\end{figure}

We first evaluate the test accuracy after compression, reported as the mean and standard deviation over ten independent trained neural networks with different seeds. 

Figure~\ref{fig:mnist_results_accuracy} shows the numerical results. These results demonstrate better performance than the pruning method shown in Fig.~\ref{fig:results_pruning}. For reference, Fig.~\ref{fig:mnist_results_accuracy} also shows the $\ell_2$-norm weight-magnitude pruning results, i.e., ``Weight magnitude pruning (L2)''; we see that the merging methods work well even with significant compression compared with the pruning results in Fig.~\ref{fig:results_pruning}. 

From Fig.~\ref{fig:mnist_results_accuracy}, we see that ``C-weight/M-logit(data)'' achieves the highest mean accuracy; the method corresponds to data-free clustering and data-dependent merging. In addition, when a large number of nodes are removed and the node-retention ratio is small, the data-free clustering approach ``C-weight'' yields better accuracy than the ``C-activation (data)'' approach. These results could be interpreted as follows.
\begin{itemize}
\item While the activation function values depend on the data, they contain only partial information. Therefore, when selecting which clusters to merge, weights that aggregate information are preferable.
\item For neuron merging, the data-dependent approach is preferable to a simple linear combination. This is because the least-squares method is effective for generating sigmoid functions with a sufficient amount of data.
\end{itemize}

It is reasonable to expect that additional information, such as the data itself, would lead to better merging results. The results shown in Fig.~\ref{fig:mnist_results_accuracy} confirm that this hypothesis is correct.

\subsubsection{One-time compression time}

\begin{figure}[tb]
  \centering
  \includegraphics[width=85mm]{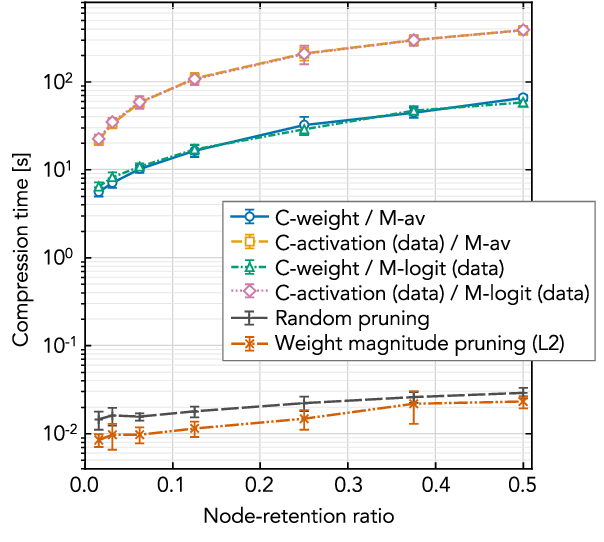}
  \caption{(Color online) One-time compression time in the MNIST dataset experiments. The horizontal axis shows the node-retention ratio in the hidden layers, and the vertical axis shows the computational time on a logarithmic scale. Symbols and error bars denote the mean and standard deviation over ten independently trained networks.}
  \label{fig:mnist_results_computatoinal_time}
\end{figure}

We here report the one-time compression time on a logarithmic scale in Fig.~\ref{fig:mnist_results_computatoinal_time}. All numerical experiments were conducted on a Apple M2 core (8 CPUs) and 8 GM of RAM. The compression time does not include evaluation time, but it does include other processing time including clustering and merging procedures. For C-activation and M-logit, the results include the time required to input data and obtain the activation function values. 

First, the node merging method takes longer than the pruning method. However, as mentioned earlier, these results refer to the computational time required for compression. Note that since all methods share the same compression architecture for a given target node-retention ratio, the number of nodes and the inference cost of the densely connected layers are identical. We perform the compression procedure only once, and the differences in compression time are not a significant issue in practice.

Second, one might consider that evaluating activation functions takes a long time, but ``C-weight/M-logit(data)'' does not. This result indicates that the $k$-means algorithm takes more time than evaluating the activation functions. This is because the size of the features in C-activation is determined by $N_{\mathrm{comp}}$, resulting in feature vectors that are larger than C-weight in the current experimental settings. Since the least-squares method is simple, we can see that it does not require much computation time.

\subsection{Training-data-free approach with randomly generated data}

In the C-activation and M-logit procedures, activation function values are determined from actual data and then used for clustering and neuron merging. Therefore, these procedures cannot be used without actual data. However, by inputting randomly generated input vectors, $\{\bm{x}_n\}$, into trained neural networks, it is possible to obtain the activation function values required for neuron merging. Hence, as a data-free method, we test the C-activation and M-logit procedures using randomly generated data.

We here randomly generate inputs as follows:
\begin{align}
  {\bm{x}}_n
  \sim \mathcal{N}(\bm{0},I_{d_0}),
  \qquad n=1,\ldots,N_{\mathrm{comp}},
  \label{eq:synthetic_inputs}
\end{align}
where $d_0$ is the input dimension.

\begin{figure}[t]
\centering
\includegraphics[width=85mm]{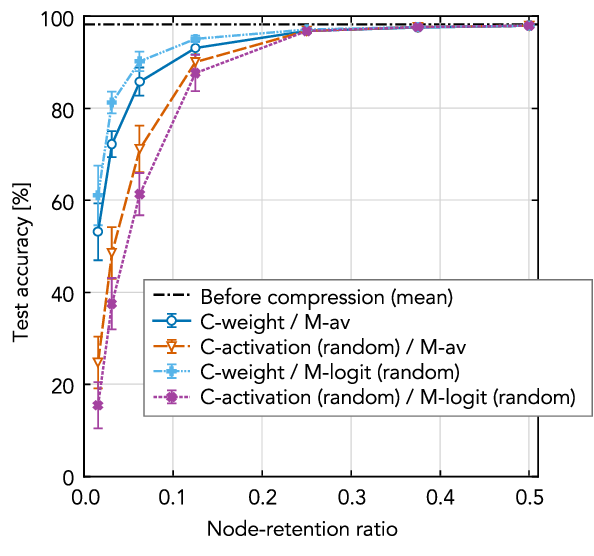}
\caption{(Color online) Comparison of test accuracy using data-free procedures in the MNIST dataset experiments. The horizontal axis shows the node-retention ratio in the hidden layers, and the vertical axis shows the accuracy on the test data. Symbols and error bars denote the mean and standard deviation over ten independently trained networks.}
  \label{fig:mnist_results_data_free}
\end{figure}

Figure~\ref{fig:mnist_results_data_free} shows the numerical results on the data-free settings. It is noteworthy that while using activation function values for clustering leads to poorer performance, using them for merging improves performance. 

As mentioned above, C-activation lacks the information of the focused neurons to some extent, leading to performance degradation. Furthermore, the performance is significantly worse than that shown in Fig.~\ref{fig:mnist_results_accuracy} when using actual data. This suggests that information about the data distribution is important. 

On the other hand, the randomly generated data is beneficial for the merging step. This is because, in the merging step, it suffices to seek sigmoid functions that approximate the input-output relationships; this is a simple least-squares problem. In our experiments, standard-normal inputs provided a sufficiently well-conditioned design matrix to obtain useful representatives, despite the distribution mismatch. Hence, the randomly generated data improved performance.

\subsection{Additional evaluation on Fashion-MNIST}
\label{subsec:fashion_results}

\begin{figure}[t]
  \centering
  \includegraphics[width=85mm]{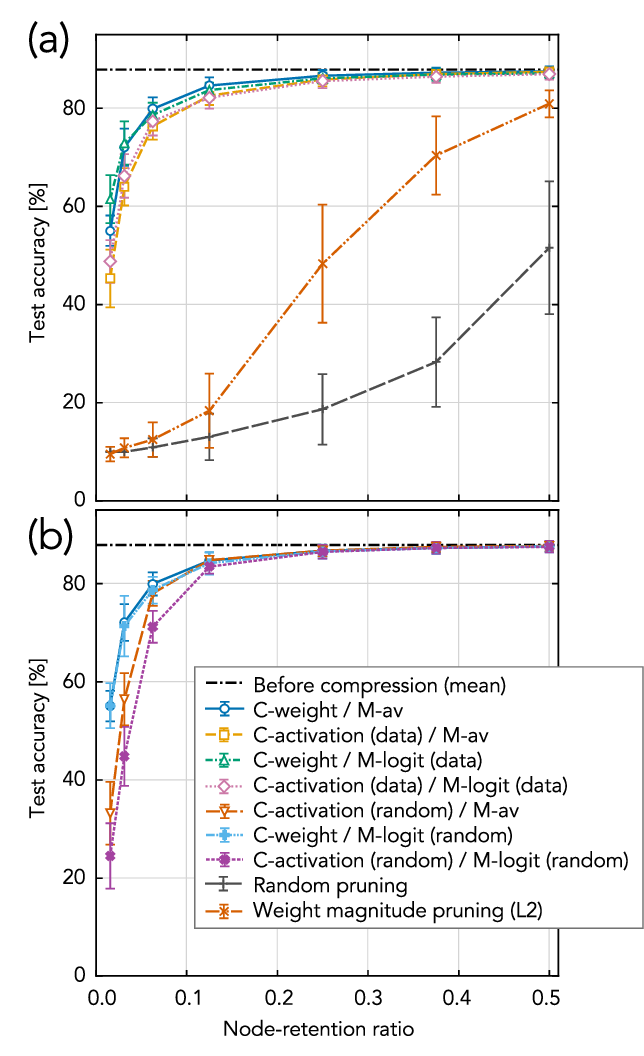}
  \caption{(Color online) Test accuracy after neuron merging in the Fashion-MNIST dataset experiments. The horizontal axis shows the node-retention ratio in the hidden layers, and the vertical axis shows the accuracy on the test data. For readability, panel (a) shows the results of the conventional pruning methods, data-free methods, and data-dependent methods, whereas panel (b) shows the training-data-free methods with randomly generated datasets; ``C-weight/M-av'' is repeated in both panels as a common reference. Symbols and error bars denote the mean and standard deviation over ten independently trained networks.
}
  \label{fig:fashion_results}
\end{figure}

Here, we show the numerical results for the Fashion-MNIST dataset \cite{Xiao2017}; the neural network architecture is the same as that used for the MNIST dataset. Since the computation time required for compression followed the same trend as the MNIST cases, we show only the test accuracy results after compression. Figure~\ref{fig:fashion_results} shows the results. These results indicate that, although the difference in accuracy is not as significant as that observed in MNIST, the trend is the same. Notably, at moderate compression rates (moderate node-retention ratios), the data-free C-weight/M-av method yields good results. However, at high compression rates (small node-retention ratios), the C-weight/M-logit method, which utilizes data in the merging, demonstrates superior performance. Furthermore, the merging method with randomly generated data also achieves sufficient accuracy.

The results are slightly less distinct since the Fashion-MNIST task is more complex than MNIST. However, the numerical findings are consistent with those of the MNIST analysis, suggesting that exploiting data information in the merging step is effective.

\section{Conclusion}
\label{sec:conclusion}

We proposed a post-training neuron-merging framework for invertible activation functions that reconstructs representative neurons by the least-squares method in the inverse-activation space. This proposal made it possible to discuss the usefulness of data in neuron merging. As a concrete instance, we focused on sigmoid networks and used the logit function without fine-tuning. 

Inverse-activation regression with weight-based clustering maintains high accuracy over a wide range of compression ratios. This result supports the usefulness of activation information for reconstructing representative neurons, while weight information remains effective for clustering. We also found that even randomly generated inputs retained substantially higher accuracy than pruning at the tested high-compression settings. It indicates that the randomly generated inputs provided sufficiently diverse activation patterns in our experiments to estimate representative sigmoid neurons. On the other hand, it has also become clear that the activation-based clustering using randomly generated inputs degraded more strongly. Because its neuron features are also determined from responses to random inputs, the mismatch between synthetic and task distributions may contribute to the larger degradation observed at high compression ratios.

The current numerical experiments are limited to fully connected sigmoid networks and relatively small datasets. Although the formulation is applicable to activation functions with usable inverses, its behavior with other invertible activation functions has not yet been evaluated. Of course, non-invertible activation functions such as ReLU require a different reconstruction strategy. In addition, the current random-input experiment uses only isotropic standard-normal inputs; other synthetic distributions may better reproduce task-relevant responses. Research that takes such practical considerations into account remains an important direction for future work. In particular, some methods, such as extreme learning machines \cite{Huang2004,Huang2006,Wang2022}, Ridgelet-based methods \cite{Murata1996,Sonoda2014,Homma2024}, and the SWIM (sampling where it matters) algorithm \cite{Bolager2023,Homma2026}, allow us to create redundant networks and train them without using backpropagation. As for the networks capable of such simple learning, it may also be important, from a practical perspective, to apply the neuron-merging method presented in this paper to reduce redundancy.

\begin{acknowledgments}
The authors thank Fumito Kimura for useful discussions. This work was financially supported in part by grant (JSPS KAKENHI Grant Number JP26K02994).
\end{acknowledgments}


%


\begin{thebibliography}{11}%

\bibitem{Strubell2019}
E. Strubell, A. Ganesh, and A. McCallum, Proc. 57th Ann. Meeting Assoc. Comp. Linguistics, Italy, 2019, p.~3645.

\bibitem{Strubell2020}
E. Strubell, A. Ganesh, and A. McCallum, Proc. AAAI Conf. Artificial Intell., U.S.A., 2020, Vol.~34, p.~13693.


\bibitem{Fernandez2025}
J. Fernandez, C. Na, V. Tiwari, Y. Bisk, S. Luccioni, and E. Strubell, Proc. 63rd Ann. Meet. Assoc. Comp. Linguistics, Austria, 2025, Vol.~1, p.~32556.

\bibitem{Caravaca2025}
F. Caravaca, SIGMETRICS Perform. Eval. Rev. \textbf{53}, 18 (2025).


\bibitem{Hinton2015}
G. Hinton, O. Vinyals, and J. Dean, NIPS Deep Learn. Workshop, arXiv:1503.02531.


\bibitem{Moslemi2024}
A. Moslemi, A. Briskina, Z. Dang, and J. Li, Mach. Learn. Appl. \textbf{18}, 100605 (2024).

\bibitem{Mansourian2025}
A.M. Mansourian, R. Ahmadi, M. Ghafouri, A.M. Babaei, E.B. Golezani, Z.Y. Ghamchi, V. Ramezanian, A. Taherian, K. Dinashi, A. Miri and S. Kasaei, Trans. Mach. Learn. Res., 2835-8856 (2025).


\bibitem{Liang2021}
T. Liang, J. Glossner, L. Wang, S. Shi, and X. Zhang, Neurocomputing \textbf{461}, 370 (2021).

\bibitem{He2024}
Y. He and L. Xiao, IEEE Trans. Pattern Anal. Mach. Intell. \textbf{46}, 2900 (2024).

\bibitem{Cheng2024}
H. Cheng, M. Zhang and J. Q. Shi, IEEE Tran. Pattern Anal. Mach. Intell. \textbf{46}, 10558 (2024).


\bibitem{Han2015}%
S. Han, J. Pool, J. Tran, and W.~J. Dally, Adv. Neur. Info. Proc. Sys. \textbf{28}, 1135 (2015).


\bibitem{Li2017}
H. Li, A. Kadav, I. Durdanovic, H. Samet, and H. P. Graf, Proc. Int. Conf. Learn. Repr., France, 2017.

\bibitem{Molchanov2017}
P. Molchanov, S. Tyree, T. Karras, T. Aila, and J. Kautz, Proc. Int. Conf. Learn. Repr., France, 2017.


\bibitem{Acharyya2021}
R. Acharyya, A. Chattoraj, B. Zhang, S. Das, and D. \v{S}tefankovi\v{c}, Proc. Mach. Learn. Res. \textbf{161}, 1988 (2021).

\bibitem{Mori2025}
F. Mori and F. Mignacco, Phys. Rev. E \textbf{112}, 045301 (2025).

\bibitem{Surkov2026}
A. Surkov, S. Koltcov, V. Ignatenko, R. Mehmood, and K. Kupitman, Physica A \textbf{681}, 131085 (2026).




\bibitem{Srinivas2015}
S. Srinivas and R. V. Babu, Proc. British Mach. Vision Conf., U.K., 2015, p~31.1.

\bibitem{Zhong2018}
G. Zhong, H. Yao, and H. Zhou, Proc. 24th Int. Conf. on Pattern Recognition, China, 2018, p.~1462.

\bibitem{Wang2020}
L.-N. Wang, W. Liu, X. Liu, G. Zhong, P. P. Roy, J. Dong, and K. Huang, Sensors \textbf{20}, 6033 (2020).

\bibitem{Emerald2025}
S. C. Emerald, and T. Vengattaraman, Int. J. Info. Tecnol. \textbf{17}, 3397 (2025).


\bibitem{Kim2020}
W. Kim, S. Kim, M. Park, and G. Jeon, Adv. Neur. Info. Proc. Sys. \textbf{33}, 585 (2020).


\bibitem{Theis2018}
L. Theis, I. Korshunova, A. Tejani, and F. Husz{\'a}r, arXiv:1801.05787.

\bibitem{He2017}
Y. He, X. Zhang, and J. Sun, Proc. IEEE Int. Conf. Comp. Vision, Italy, 2017, p.~1398.



\bibitem{Lecun1998}
Y. LeCun, L. Bottou, Y. Bengio, and P. Haffner, Proc. of the IEEE, 1998, Vol.~86, p.~2278.

\bibitem{scikit-learn}
\url{https://scikit-learn.org/}

\bibitem{Xiao2017}
H. Xiao, K. Rasul, and R. Vollgraf, arXiv:1708.07747.




\bibitem{Huang2004}
G.-B. Huang, Q.-Y. Zhu, and C.-K. Siew, Proc. 2004 IEEE Int. Joint Conf. Neural Networks, Hungary, 2004, p.~985.

\bibitem{Huang2006}
G.-B. Huang, Q.-Y. Zhu, and C.-K. Siew, Neurocomputing \textbf{70}, 489 (2006).

\bibitem{Wang2022}
J. Wang, S. Lu, S.-H. Wang, and Y.-D. Zhang, Multimed. Tools Appl. \textbf{81}, 41611 (2022).


  
\bibitem{Murata1996}
N. Murata, Neural Networks \textbf{9}, 947 (1996).

\bibitem{Sonoda2014}
S. Sonoda and N. Murata, Lect. Notes Comp. Sci. \textbf{8681}, 539 (2014).

\bibitem{Homma2024}
H. Homma and J. Ohkubo, J. Phys. Soc. Jpn. \textbf{93}, 124001 (2024).

\bibitem{Bolager2023}
E. L. Bolager, I. Burak, C. Datar, Q. Sun, and F. Dietrich, Adv. Neur. Info. Proc. Sys. \textbf{36}, 63075 (2023).

\bibitem{Homma2026}
H. Homma and J. Ohkubo, J. Phys. Soc. Jpn. \textbf{95}, 044002 (2026).

\end{thebibliography}

\end{document}